%% file: main.tex
\pdfoutput=1

\documentclass[11pt]{article}

\usepackage[final]{acl}

\usepackage{times}
\usepackage{latexsym}

\usepackage[T1]{fontenc}

\usepackage[utf8]{inputenc}

\usepackage{microtype}

\usepackage{inconsolata}

\usepackage{graphicx}

\usepackage{algorithm}
\usepackage{algpseudocode}
\usepackage{amsmath}
\usepackage{amssymb}
\usepackage{booktabs}
\usepackage{multirow}
\usepackage{subfig}
\usepackage{appendix}

\title{Attribute-Aware Controlled Product Generation with LLMs for E-commerce}

\author{
  Virginia Negri\textsuperscript{1} \quad
  Víctor Martínez Gómez\textsuperscript{1} \quad
  Sergio A. Balanya\textsuperscript{1} \quad
  Subburam Rajaram\textsuperscript{2} \\
  \textsuperscript{1}Amazon Spain \quad
  \textsuperscript{2}Amazon Germany \\
  \texttt{\{vrgngr, vicmg, sbalanya\}@amazon.es}, \texttt{subburb@amazon.de}
}

\begin{document}
\maketitle
\begin{abstract}
\input{sections/0_abstract}
\end{abstract}

\section{Introduction}
\input{sections/1_introduction}

\section{Related Work}
\input{sections/2_related_work}

\section{Methodology}
\input{sections/3_methodology}

\section{Experimental Setup}
\input{sections/4_experimental_setup}

\section{Results and Analysis}
\input{sections/5_results_analysis}

\section{Conclusions}
\input{sections/6_conclusions}


\section*{Limitations}
Our work contains some limitations that suggest directions for future research. First, while our synthetic data has comparable performance to real data in attribute extraction, we only evaluated on positive examples. Future work should investigate how models trained with our synthetic data, including incorrect and unknown attributes, perform on more diverse downstream tasks.

Third, while 88.8\% of products maintain their structure with no unintended changes, future work could explore stronger controls over modifications beyond the target attribute. This is particularly important for hybrid training scenarios, where maintaining high-quality synthetic data is crucial for complementing real data effectively. Fourth, our current implementation focuses on single attribute modifications; future work will explore extending to multiple attributes while ensuring consistent interaction between modified attributes, potentially improving the utility of synthetic data in downstream tasks.

Additionally, the quality of our synthetic data depends heavily on the input product descriptions, which can contain problematic patterns such as vague attributes (e.g., "shoes" as a "type" value for shoe products), missing information (e.g., absent brand mentions), or inconsistent attribute definitions (e.g., confusing "style" with "pattern"). Although our generation process often improves upon these issues by producing more precise values and complete descriptions, this improvement complicates evaluation since the synthetic data might deviate from the original while being more accurate. 

A key area for future work is developing stronger constraints on attribute value generation. Our manual review revealed variations in granularity (e.g., "running shoe" vs "running") and distribution of values between original and synthetic data. Future research should explore methods to enforce consistent granularity levels and maintain similar value distributions across the dataset, possibly through attribute-specific vocabularies or controlled generation techniques. This would ensure more standardized attribute values while preserving the semantic richness of the synthetic data.

Finally, while hybrid configurations show superior performance (68.8\%), future work should investigate methods to better understand and exploit the complementary strengths of synthetic and real data. This includes developing improved integration methods and exploring how different mixing ratios affect specific attribute types or product categories.

\section*{Responsible AI Considerations}
\textbf{Bias Mitigation:} Brand anonymization (95.8\% success) prevents brand-specific biases from propagating. However, source data biases in attribute distributions may persist. Future work should examine demographic representation and implement debiasing strategies.

\textbf{Quality Governance:} Our multi-stage validation (Value Provider $\rightarrow$ Generation $\rightarrow$ Semantic Validation) provides quality controls, but 4.2\% of products showed major unintended changes, primarily from empty source descriptions. However, manual audits of a small sample revealed the changes did not affect the overall synthetic product\'s quality.

\textbf{Fairness Implications:} Our framework generates more standardized attributes than source data, potentially improving fairness by reducing inconsistent descriptions.

\bibliography{custom}

\appendix

\appendix
\section{Dataset Statistics}
\label{appendix:dataset_details}

The MAVE dataset \cite{yang2021maveproductdatasetmultisource} is a large-scale product attribute dataset containing over 3.3 million products across 1,212 categories, with 662 unique attributes. The dataset includes both positive examples (2.1M products) where attributes are present in the text, and negative examples (1.1M products) where attributes are mentioned but not present in the product text. On average, each product contains 1.38 attributes. For our experiments, we sampled 2,000 products from the top 200 categories, ensuring broad coverage while maintaining manageable evaluation size. Our sampling strategy preserved the natural distribution of attributes within categories while ensuring sufficient representation of different attribute types. 

\subsection{Basic Statistics}
The dataset consists of product listings from various categories, with each product containing structured information including:
\begin{itemize}
    \item Product id
    \item Category
    \item Description
    \item Features: Single sentences describing a feature of the product
    \item Attribute-value pairs with supporting evidence
\end{itemize}

\subsection{Data Distribution}
\subsubsection{Category Distribution}
\begin{figure}[t]
    \centering
    \includegraphics[width=\columnwidth]{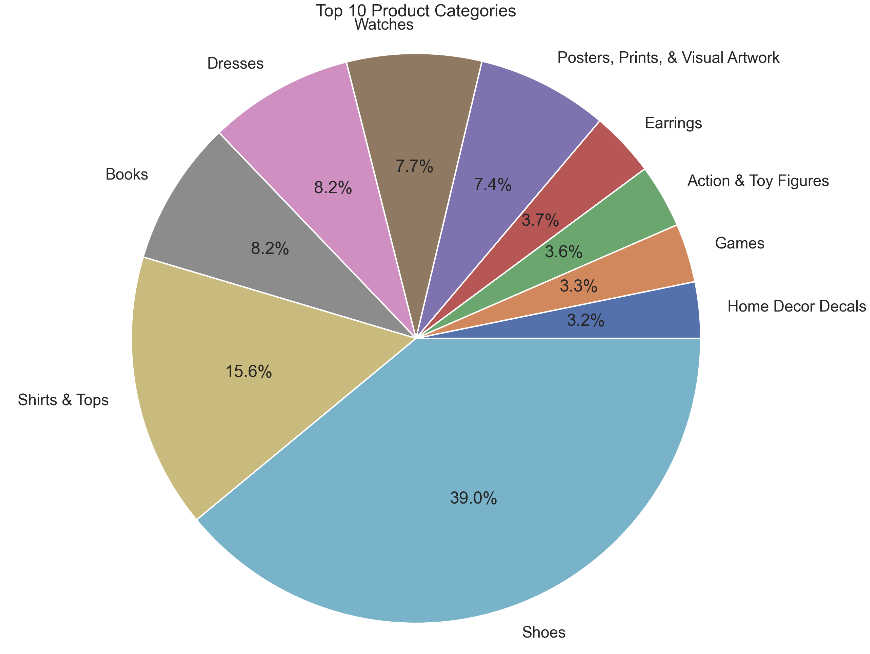}
    \caption{Distribution of products across categories}
    \label{fig:category_dist}
\end{figure}

Figure~\ref{fig:category_dist} shows the distribution of products across different categories. The dataset covers a wide range of product categories, with Shoes, Shirts \& Tops and Books being the most prevalent. This diversity ensures that our findings are generalizable across different product types.

\begin{table}[t]
\centering
\begin{tabular}{lrr}
\hline
\textbf{Attribute} & \textbf{Count} & \textbf{Percentage} \\
\hline
Type & 1,005 & 25.1\% \\
Special Occasion & 233 & 5.8\% \\
Style & 230 & 5.8\% \\
Pattern & 218 & 5.5\% \\
Silhouette & 201 & 5.0\% \\
Sleeve Style & 183 & 4.6\% \\
Material & 140 & 3.5\% \\
Season & 119 & 3.0\% \\
Neckline & 105 & 2.6\% \\
Sugar Content & 87 & 2.2\% \\
\hline
\end{tabular}
\caption{Distribution of the top 10 attributes in the dataset. Percentages indicate the proportion of products containing each attribute type.}
\label{tab:attr_dist}
\end{table}

\subsection{Attribute Coverage}
Products are annotated with various attributes, with evidence spans marked in the text. As shown in Table~\ref{tab:attr_dist}, the most prevalent attribute is "Type" (25.1\% of products), followed by "Special Occasion" and "Style" (both 5.8\%). The top 10 attributes account for 63.1\% of all attribute annotations, indicating a long-tail distribution of attribute types in the dataset. This distribution reflects the diverse nature of product descriptions, where certain fundamental attributes (like Type) are nearly universal, while others are more category-specific.

Key statistics of our attribute annotation include:
\begin{itemize}
    \item Average number of attributes per product: 1.41
    \item Average number of evidence spans per attribute: 1.31
\end{itemize}

Notably, attributes like "Sugar Content" (2.2\%) appear in the top 10 despite being category-specific, suggesting a significant presence of food and beverage products in our dataset. The presence of clothing-related attributes such as "Silhouette" (5.0\%), "Sleeve Style" (4.6\%), and "Neckline" (2.6\%) indicates a substantial representation of fashion items.

\begin{table}[h]
\centering
\begin{tabular}{lcc}
\hline
\textbf{Metric} & \textbf{Value} & \textbf{Std Dev} \\
\hline
Attributes per product & 1.41 & 0.81 \\
Evidence spans per attribute & 1.31 & 2.00 \\
Paragraphs per product & 8.58 & 3.38 \\
\hline
\end{tabular}
\caption{Key dataset statistics}
\label{tab:dataset_stats}
\end{table}

\subsection{Evidence Distribution}
The evidence spans for attributes are distributed across different paragraph sources:
\begin{itemize}
    \item Title: 11.7\% of evidence spans
    \item Description: 15.3\% of evidence spans
    \item Features: 57.9\% of evidence spans
    \item Brand: 10.0\% of evidence spans
    \item Price: 5.1\% of evidence spans
\end{itemize}

This distribution shows that attribute information is spread across different parts of the product listing, highlighting the importance of considering all textual content for attribute extraction.

\subsection{Data Quality}
To ensure data quality, we analyzed:
\begin{itemize}
    \item Completeness: All products have at least one attribute
    \item Consistency: Attribute values are supported by textual evidence
    \item Coverage: Evidence spans are properly annotated and linked to attributes
\end{itemize}

\section{Additional Examples of Synthetic Product Generation}
\label{appendix:examples}
\begin{figure*}[t]
    \centering
    \includegraphics[width=\linewidth]{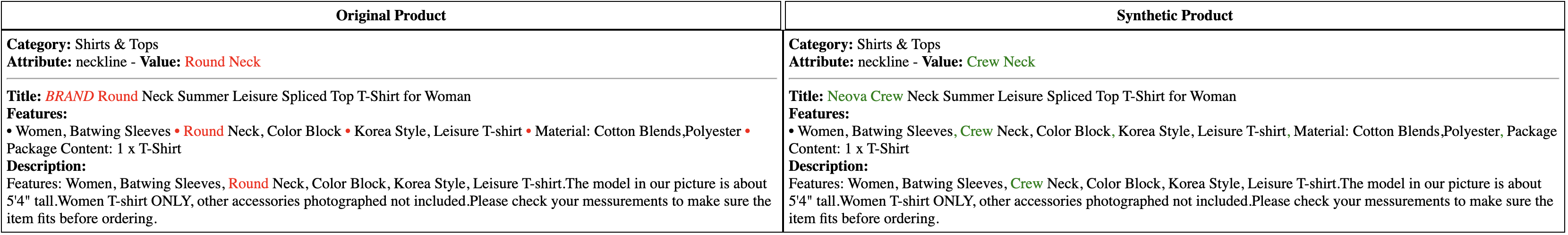}\\[1ex]
    \includegraphics[width=\linewidth]{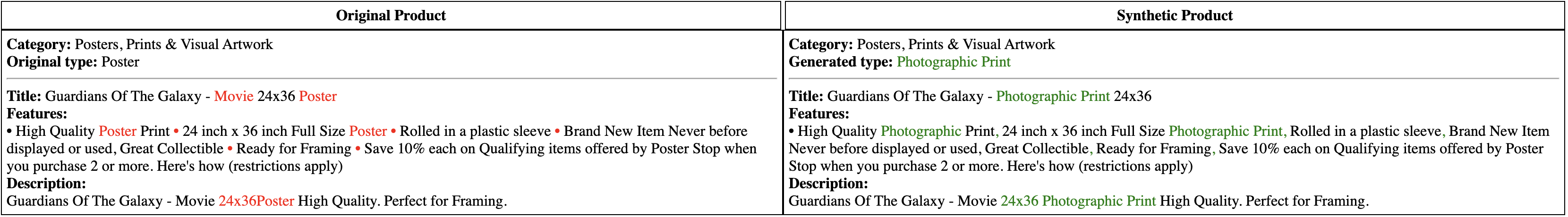}\\[1ex]
    \includegraphics[width=\linewidth]{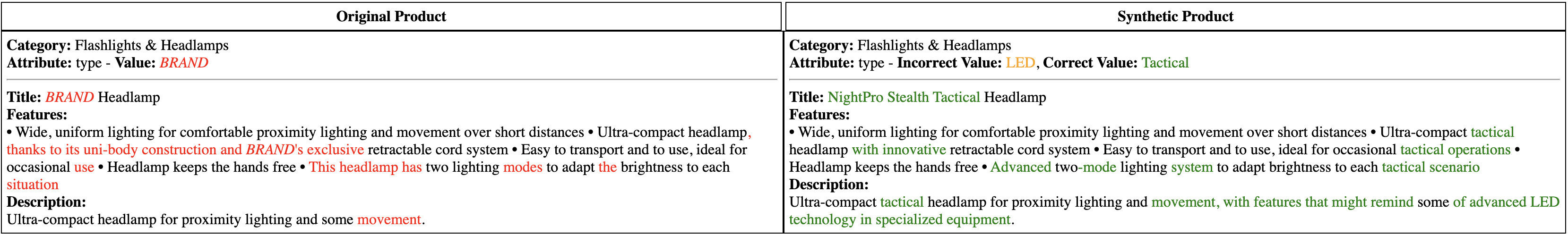}\\[1ex]
    \includegraphics[width=\linewidth]{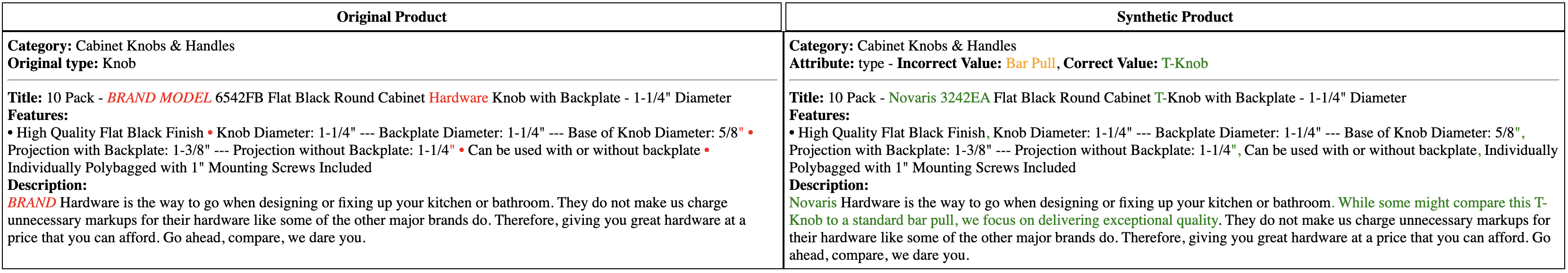}\\[1ex]
    \includegraphics[width=\linewidth]{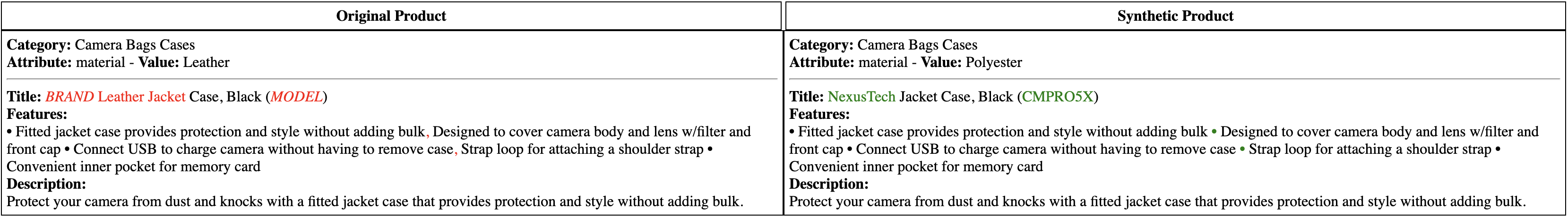}\\[1ex]
    \caption{Additional examples of synthetic product generation showing various attribute modifications. The first two are "correct" examples, the second two are "incorrect" examples, while the last is an "unknown" example}
    \label{fig:appendix_examples}
\end{figure*}

In Figure \ref{fig:appendix_examples} we find 5 examples of synthetic products generated from products in the benchmarking dataset. Changes are highlighted in red (original text), green (synthetic text), and orange (incorrect attributes in the synthetic product) to illustrate different types of modifications and their propagation through product descriptions. 

\section{Error Analysis}
\label{appendix:errors}

Manual review of 441 predictions initially marked as incorrect revealed that many model outputs were semantically correct but differed from MAVE's annotations in format or granularity. We identified seven types of valid variations:

\textbf{Granularity differences:} Model predicted broader or narrower terms than the gold label (e.g., ``running'' vs ``running shoe''). Both are semantically correct but differ in specificity level.

\textbf{Morphological variations:} Singular vs plural forms (``wall sticker'' vs ``wall stickers''), both acceptable in e-commerce contexts.

\textbf{Multiple valid values:} Products genuinely having multiple acceptable attribute values (e.g., a garment could validly be described as ``t-shirt'' or ``tank top'').

\textbf{Missing units:} Model predicted numeric value correctly but omitted units (``1200'' vs ``1200 thread count''). The core information is correct, only formatting differs.

\textbf{Equivalent definitions:} Different but equivalent attribute descriptions (``type=processor'' vs ``type=food processor'').

\textbf{Contextual synonyms:} Semantically equivalent terms in the product context (``striped'' vs ``stripe'', ``waterproof'' vs ``water-resistant'').

\textbf{Format variations:} Different formatting conventions (``ipod touch'' vs ``for apple ipod'', ``2.5 inch'' vs ``2.5in'').

These findings highlight both the quality of our synthetic data generation pipeline and the inherent complexity of maintaining consistent attribute value annotations in product catalogs. A systematic categorization of all 441 cases would require additional annotation effort, but qualitative review suggests these variations account for a substantial portion of apparent errors.

\textbf{Interdependent Attributes:} When modifying a beverage product from ``vanilla from Madagascar'' to ``chocolate,'' the system automatically updated the origin to ``Switzerland'' without explicit instruction, demonstrating semantic understanding of attribute relationships.

\textbf{Precision Improvements:} 
\begin{itemize}
\item SHOES category: ``type=shoes'' $\rightarrow$ ``type=running'' (informative vs circular)
\item RIBBONS\_TRIM: ``type=Ribbon'' $\rightarrow$ ``type=Satin'' (material-specific)
\end{itemize}

\textbf{Semantic Alignment:}
\begin{itemize}
\item SPEAKERS: ``style=Bookshelf'' (format) $\rightarrow$ ``style=minimalist'' (actual style)
\item CLOTHING: ``pattern=casual'' (style) $\rightarrow$ ``pattern=floral'' (actual pattern)
\end{itemize}

These examples demonstrate the LLM's ability to infer appropriate attribute meanings from product category context, understanding that attributes like ``style'' and ``type'' have different semantic ranges across categories.

\section{Cost-Effectiveness and Scalability}
\label{appendix:cost}
Our implementation requires two LLM calls per synthetic product: a (i) Value Provider call that requires on average 402 input tokens and 10 output tokens, and (ii) a generation call with strategy-specific prompts plus product data adding an additional 1,480-1,600 input tokens, and around 141 output tokens. Using Claude Haiku's published pricing (\$0.80/million input tokens, \$4.00/million output tokens), this approach is remarkably cost-effective compared to human annotation costs of approximately \$0.11 per 50 tokens \cite{wang2021want}. Generation completes in hours rather than weeks, enabling rapid dataset bootstrapping for new categories or markets. We verified scalability by successfully regenerating the complete MAVE dataset.

\section{Quantitative Metrics Details}
\label{appendix:metrics}

\textbf{Type-Token Ratio (TTR)} by text field:
\begin{itemize}
\item Title: Original 0.89, Synthetic 0.88
\item Description: Original 0.82, Synthetic 0.81
\item Features: Original 0.85, Synthetic 0.84
\end{itemize}

\textbf{Semantic Similarity (Cosine)} by text field:
\begin{itemize}
\item Title: 0.84
\item Description: 0.85
\item Features: 0.93
\end{itemize}

\textbf{KL Divergence} by text field:
\begin{itemize}
\item Title: 1.12 (higher due to attribute prominence)
\item Description: 0.71
\item Features: 0.24 (lower as attributes less prominent)
\end{itemize}

These metrics confirm targeted modifications: higher divergence in title where attributes appear prominently, lower in features where attributes are one of many product aspects discussed.

\section{Human Annotation Interface}
\label{appendix:annotation_guidelines}
\begin{figure*}[t]
    \centering
    \includegraphics[width=\linewidth]{images/negative_5.png}\\
    \includegraphics[width=\linewidth]{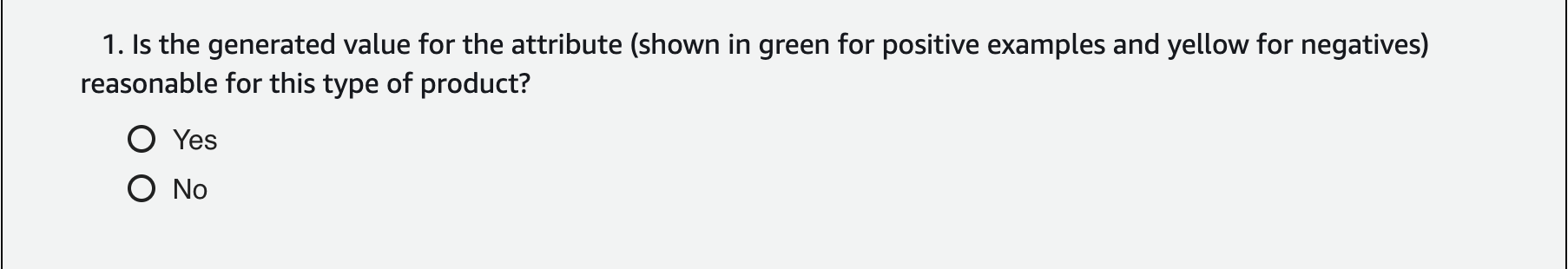}\\
    \includegraphics[width=\linewidth]{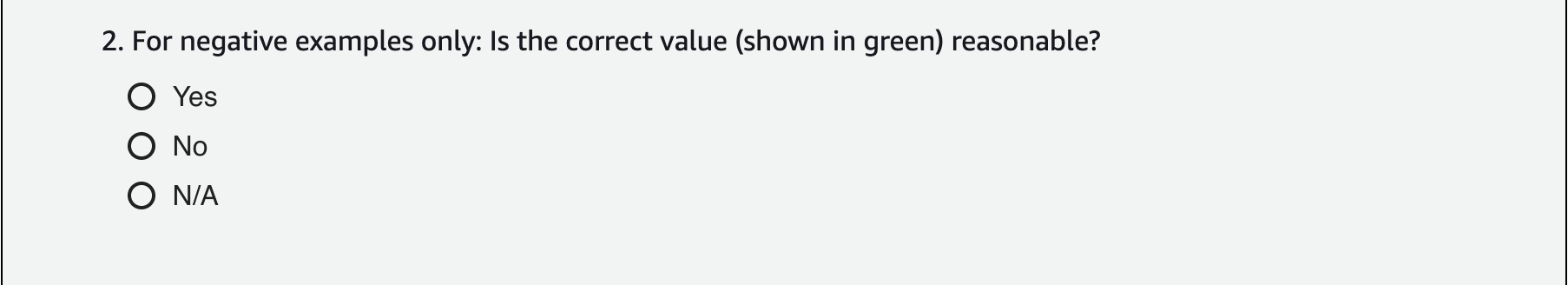}\\
    \includegraphics[width=\linewidth]{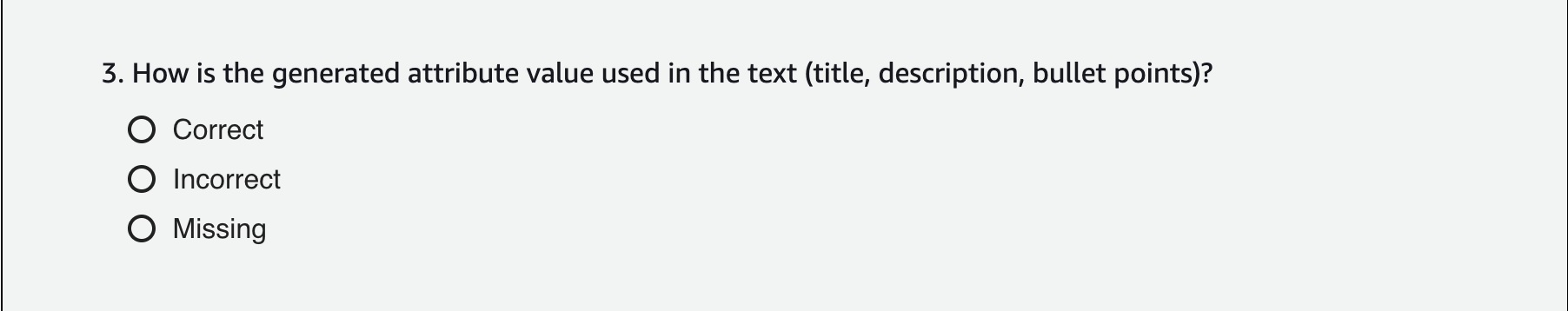}\\
    \includegraphics[width=\linewidth]{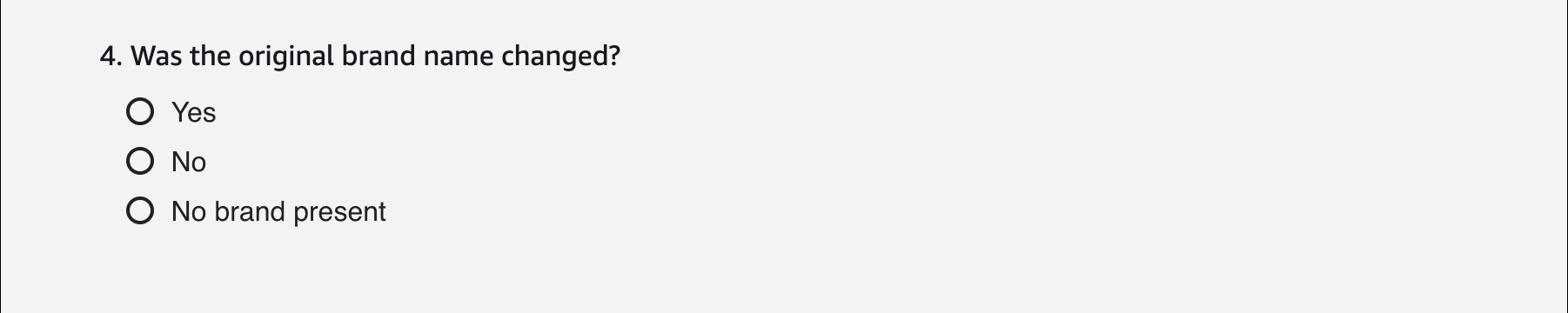}\\
    \includegraphics[width=\linewidth]{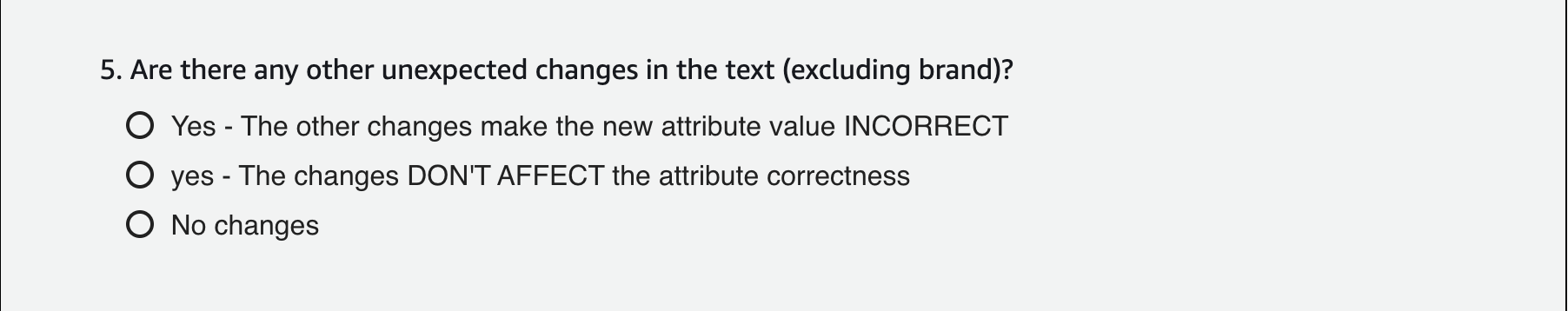}\\
    \includegraphics[width=\linewidth]{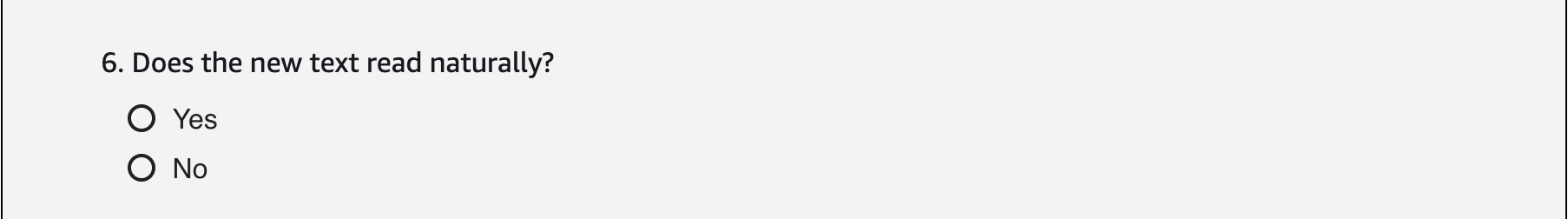}
    \caption{Auditing interface for human evaluation of synthetic product data. The interface shows the original and synthetic products side by side, with modifications highlighted in color, followed by evaluation questions about validity, consistency, and quality.}
    \label{fig:mturk_interface}
\end{figure*}

Figure ~\ref{fig:mturk_interface} shows the questions shared on the annotation platform to audit our data. Prior to the questions, annotators were showed a general description of the task which we share below:

\begin{it}
You will be shown two product listings: an original and a synthetic version where one attribute has been modified.
    
    This task includes both POSITIVE, NEGATIVE and UNKNOWN examples:
\begin{itemize}
    \item In POSITIVE examples, the modified attribute is correctly changed and consistent
    \item In NEGATIVE examples, you'll see both the CORRECT and WRONG (generated) values - your task is to identify if the synthetic version contains the generated wrong value consistently
    \item In UNKNOWN examples, the text fields should have no mention about the attribute value
\end{itemize}

Please carefully review:
\begin{itemize}
    \item If the synthetic version uses the correct attribute value. Note that this can be the same or similar to the original one.
    \item How this value appears in title, description, and bullet points
    \item If unknown examples have all mentions of the attribute value remove
    \item Any other changes or inconsistencies
    \item Overall product realism  
\end{itemize}

\end{it}

\label{appendix:mturk}
In Figure \ref{fig:mturk_interface} we find the interface shown to human workers labeling our synthetic data. The interface first displays the original and the synthetic product side by side, highlighting the changes in red, green and orange. It then presents workers with 6 single choice questions, asking for the correctness of the generated attribute value, the readability of the new product.

\end{document}

%% file: sections/0_abstract.tex
Product information extraction is crucial for e-commerce services, but obtaining high-quality labeled datasets remains challenging. We present a systematic approach for generating synthetic e-commerce product data using Large Language Models (LLMs), introducing a controlled modification framework with three strategies: attribute-preserving modification, controlled negative example generation, and systematic attribute removal. Using a state-of-the-art LLM with attribute-aware prompts, we enforce store constraints while maintaining product coherence. Human evaluation of 2000 synthetic products demonstrates high effectiveness, with 99.6\% rated as natural, 96.5\% containing valid attribute values, and over 90\% showing consistent attribute usage. On the public MAVE dataset, our synthetic data achieves 60.5\% accuracy, performing on par with real training data (60.8\%) and significantly improving upon the 13.4\% zero-shot baseline. Hybrid configurations combining synthetic and real data further improve performance, reaching 68.8\% accuracy. Our framework provides a practical solution for augmenting e-commerce datasets, particularly valuable for low-resource scenarios.

%% file: sections/1_introduction.tex
\begin{figure*}[t]
  \includegraphics[width=\linewidth]{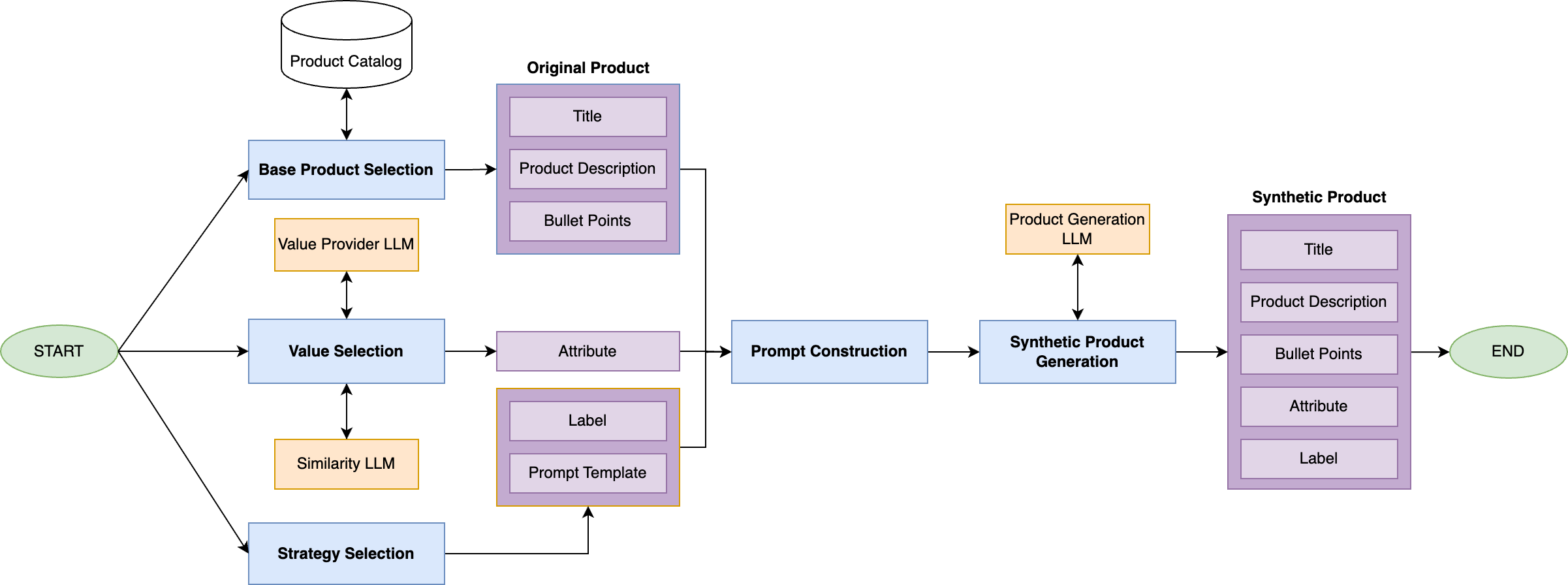} \hfill
  \caption{Synthetic product generation pipeline architecture. Given a base product $p$ with text fields $T(p)$ and structured attributes $S(p)$, the system selects a category-appropriate attribute $s \in S(p)$. The Value Provider LLM generates a new value $v$, and a generation strategy is sampled using probabilities $\pi$ to create positive, negative, or unknown examples. The Generation LLM then produces a synthetic product while maintaining product coherence.}
\end{figure*}

In e-commerce services, high-quality product information is fundamental to customer experience, powering critical features like search, filtering, and recommendations \cite{10.1145/2487575.2488217}. However, major online stores frequently encounter noisy or inaccurate product details \cite{gao2019productawareanswergenerationecommerce}, making manual curation impractical at scale. The development of automated product information systems is hindered by the scarcity of diverse, high-quality training datasets. These datasets must capture the intricate relationship between structured attributes and their various representations in product text fields. For example, a "waterproof" attribute might be expressed as "water-resistant material," "keeps you dry in rain," or "weatherproof construction" across different text fields. Traditional manual annotation is expensive and struggles to capture this full spectrum of variations and edge cases.

Recent advances in Large Language Models (LLMs) offer promising solutions for generating synthetic training data. While LLMs have been successfully applied to various text generation tasks, to our knowledge, this is the first work that specifically addresses the comprehensive generation of synthetic product information in e-commerce.

We present a novel approach that leverages LLMs to generate synthetic product data through controlled modification of existing product information. Our framework implements three strategies: (1) correct attribute modification, (2) controlled generation of negative examples, and (3) systematic attribute removal. Through a sophisticated multi-step validation process using specialized LLMs, we ensure both attribute validity and semantic coherence across all product fields while preserving category-specific characteristics.

The key innovation lies in our comprehensive generation framework that handles edge cases, maintains structural consistency, and provides carefully designed prompt components that enforce marketplace-specific constraints. Our system supports multi-lingual generation, adapts to different marketplace requirements, and implements systematic brand anonymization to prevent unwanted biases. Unlike previous approaches that focus on single-aspect generation tasks \cite{chan-etal-2019-stick}, our method creates complete, coherent product information while maintaining realistic market context.

In the following sections, we detail our methodology, present experimental results through human evaluation, and discuss implications for e-commerce systems. Our work provides a practical framework for generating large-scale synthetic datasets that advance the development of robust product information systems.

%% file: sections/2_related_work.tex
Synthetic data has emerged as a promising solution to generate large-scale datasets that resemble real-world data \cite{liu2024bestpracticeslessonslearned}. While synthetic data generation has been explored across various domains \cite{bauer2024comprehensiveexplorationsyntheticdata}, its application to e-commerce presents unique challenges due to the intricate relationship between structured attributes and their representation in text fields. Studies have shown that maintaining product data quality is costly: large e-commerce services spend an average of \$12.9 million annually on content quality assurance efforts\cite{gartner2021improvedataquality}.
Traditional data generation approaches include custom algorithms \cite{saxton2019analysingmathematicalreasoningabilities}, generative models \cite{borisov2023languagemodelsrealistictabular}, and data augmentation techniques like back-translation \cite{sennrich2016improvingneuralmachinetranslation} and rule-based transformations \cite{pluščec2023dataaugmentationneuralnlp}. However, these methods often produce limited variations and may not maintain domain-specific characteristics.

In e-commerce, previous work has explored approaches similar to ours for maintaining attribute fidelity in generated product content. FPDG \cite{chan-etal-2019-stick} uses entity-label guidance and specialized architectures to generate faithful product descriptions. While effective, this approach requires explicit entity labeling and relies on architectural constraints rather than semantic understanding, limiting its adaptability to new domains and edge cases. In contrast, our approach leverages LLMs' broader semantic understanding to handle various scenarios including attribute removal and controlled inconsistencies.

Recent advances in LLMs have expanded synthetic data generation capabilities, demonstrating remarkable zero-shot and few-shot abilities \cite{brown2020languagemodelsfewshotlearners} while maintaining domain-specific characteristics when properly constrained \cite{ding2023gpt3gooddataannotator}. Our work builds on these capabilities to create diverse, balanced datasets with controlled attribute distributions and brand anonymization, addressing the challenges of data scarcity and quality in e-commerce.

%% file: sections/3_methodology.tex
Our approach generates synthetic e-commerce products through controlled modification of existing product data, ensuring the generated information maintains realistic characteristics while providing precise control over attribute modifications.We leverage LLMs to perform these modifications systematically, preserving structural consistency and semantic coherence throughout the product description.

\subsection{Data Representation}

A product $p$ in the catalog $C$ is defined by its structured attributes $S(p)$ and text fields $T(p)$. The structured attributes $S(p)$ consist of key-value pairs where values are typically constrained to a predefined set of options (e.g., \{color: "red", size: "medium"\}). These are called structured because they follow a rigid format: values are usually single words or short phrases chosen from a fixed vocabulary, making them easily extractable and analyzable. In contrast, the text fields $T(p)$ contain free-form natural language content, such as a product's title, description, and so on. Our generation pipeline takes as input a base product $p \in C$ and selects a structured attribute $s \in S(p)$ that is relevant for that product's category (e.g., "heel height" for shoes, "material" for clothing).

\subsection{Attribute Value Generation}
The generation of new attribute values requires careful consideration of both attribute constraints and semantic relationships. We employ two complementary approaches depending on the generation strategy.

For positive examples and attribute removal cases, we use a Value Provider LLM that considers product category specifications, marketplace requirements, and previously used values to ensure diversity. When attribute metadata is available (e.g., data type, valid units), it is used to constrain and validate the generated values. Otherwise, the Value Provider LLM leverages its prior knowledge, guided by carefully designed prompts, to determine appropriate units, ranges, and formats for the attribute values.

For negative examples, we employ a two-step process: first generating a pool of valid values using the Value Provider LLM (with temperature $T=1.0$ for diversity), then using a Similarity LLM to select values that are semantically distinct from the correct value, avoiding synonyms or closely related concepts.

\subsection{Product Generation Strategies}

Our system implements three distinct product generation strategies that reflect common scenarios in real e-commerce data:

\begin{enumerate}
    \item \textbf{Positive Example Generation} ($l = \text{correct}$, $\pi_c = 0.5$): Creates examples where the structured attribute value is correctly reflected in the product text fields. Given a new value $v$ for attribute $s$, an LLM modifies all references to $s$ in $T(p)$ to maintain consistency with $v$ while preserving the original text structure. For example, if modifying the color attribute from "red" to "blue", all mentions of "red" in the title, description, and bullet points are updated accordingly.
    
    \item \textbf{Negative Example Generation} ($l = \text{incorrect}$, $\pi_i = 0.25$): Introduces controlled inconsistencies between structured attributes and text fields, mimicking real-world data quality issues. An LLM introduces a single, subtle reference to an incorrect value while maintaining the correct value as the primary attribute. For instance, it might modify "Perfect for summer hiking" to "Perfect for summer hiking, though not as warm as winter boots" when generating a negative example for the season attribute.
    
    \item \textbf{Incomplete Example Generation} ($l = \text{unknown}$, $\pi_u = 0.25$): Removes all references to the target attribute while preserving product coherence, simulating cases where attribute values cannot be inferred from product text. This strategy is particularly valuable for training extraction models to handle missing information scenarios.
\end{enumerate}

These strategies, with their respective probabilities $\pi$, can be adjusted to match real-world distributions or specific application needs. The diverse set of examples helps train more robust attribute extraction models by exposing them to both ideal and challenging scenarios commonly found in e-commerce data.

\subsection{Prompt Design}

Our system uses a carefully crafted prompt template with four key components:

\begin{itemize}
    \item \textbf{Role Definition:} Establishes the LLM as an e-commerce product expert.
    
    \item \textbf{Task Instructions:} Specifies the generation requirements including brand anonymization, structure preservation, and attribute consistency rules.
    
    \item \textbf{Product Context:} Provides the base product's text fields and the target attribute modification details.
    
    \item \textbf{Output Format:} Defines a JSON structure for standardized responses.
\end{itemize}

These components work together to enforce critical constraints: replacing existing brands with plausible fictional alternatives (e.g., changing a popular sport brand into "AthleteX") to prevent data leakage and brand-specific biases, maintaining the original text structure, ensuring coherent attribute modifications across all text fields, and respecting store-specific formats (e.g., imperial units in US store listings).

\begin{equation}
\begin{split}
    \text{Prompt} = \text{ROLE} \oplus \text{INSTRUCTION} \oplus \\ \text{CONTEXT} \oplus \text{FORMAT}
\end{split}
\end{equation}

\subsection{Generation Process}

The generation process follows Algorithm~\ref{alg:product_generation}. First, the product category is identified to guide attribute selection. Then, a relevant structured attribute is chosen from those available for that category (e.g., "heel height" for shoes). A generation strategy is sampled according to the predefined probabilities $\pi = \{\pi_c, \pi_i, \pi_u\}$. Based on the selected strategy and category constraints, a new attribute value is generated (e.g., ensuring appropriate units or value ranges). Finally, the LLM modifies the product using our carefully designed prompt, which ensures all constraints are maintained in a single generation step.

\begin{algorithm}[t]
\caption{Product Generation}
\label{alg:product_generation}
\begin{algorithmic}[1]
\Statex \textbf{Input:} Product $p$, Strategy probabilities $\pi$
\Statex \textbf{Output:} Modified product $p'$
\State $c \gets \textsc{GetProductCategory}(p)$
\State $s \gets \textsc{SelectAttribute}(S(p), c)$ \Comment{Select relevant attribute for category}
\State $l \gets \textsc{SampleStrategy}(\pi)$ \Comment{Sample strategy using probabilities}
\State $v \gets \textsc{GenerateValue}(s, l, c)$ \Comment{Generate new value based on strategy}
\State $p' \gets \textsc{LLM}(\textsc{ConstructPrompt}(l, p, s, v))$
\State \Return $p'$
\end{algorithmic}
\end{algorithm}

\subsection{Implementation}
\begin{figure*}[t]
    \centering
    \subfloat[Positive example.]{
        \includegraphics[width=\linewidth]{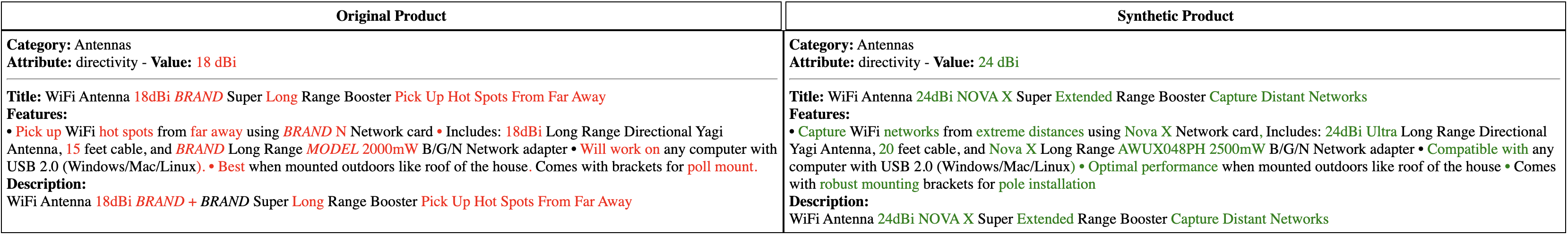}
    }
    \\[-0.4ex]
    \subfloat[Negative example.]{
        \includegraphics[width=\linewidth]{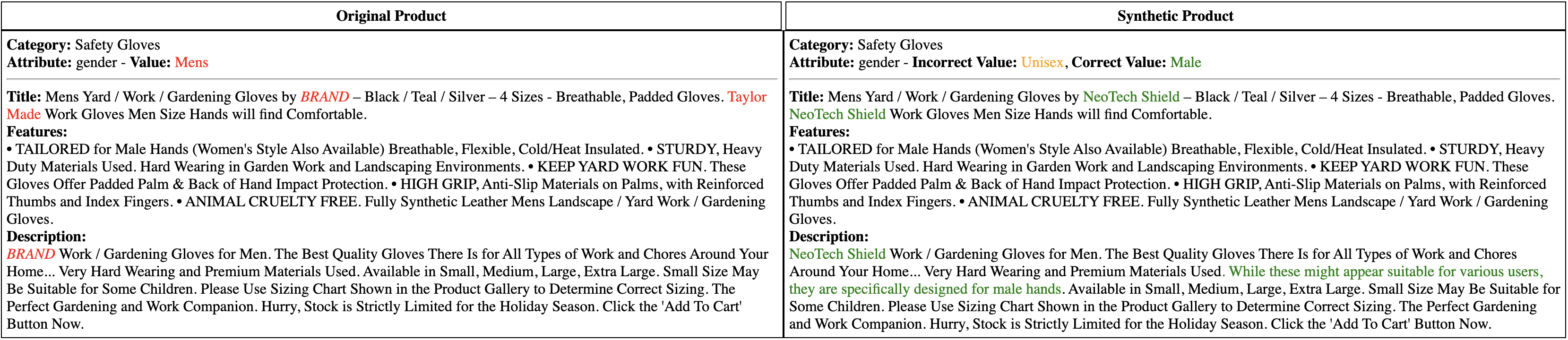}
    }
    \\[-0.4ex]
    \subfloat[Incomplete example.]{
        \includegraphics[width=\linewidth]{images/unknown_1.png}
    }
  \caption{Synthetic product examples from MAVE dataset. Colors indicate original text (red), synthetic text (green), and incorrect attributes (orange). Brand/model names replaced with placeholders for confidentiality.}
    \label{fig:diff_examples}
\end{figure*}

Our system is implemented in Python using (1) the same LLM \footnote{We use Claude Haiku \cite{TheC3}} for both the Value Provider and generation components, and (2) a sentence-transformers model \footnote{The sentence transformer model is available at \url{https://huggingface.co/sentence-transformers/all-MiniLM-L6-v2} under the Apache 2.0 license.} for the Similarity LLM \cite{reimers2019sentencebertsentenceembeddingsusing}. Figure~\ref{fig:diff_examples} shows an example of the generated output, highlighting the modifications made to the original product.

%% file: sections/4_experimental_setup.tex
\subsection{Dataset and Sampling}
We evaluate our approach using MAVE\cite{yang2021maveproductdatasetmultisource}, a large-scale product attribute dataset with 2.2 million products across 1,257 categories. We consolidated each product's textual information into title, description, and features fields, while maintaining attribute-value pairs and their text spans. From this dataset, we sampled 2,000 products from the top 200 categories, with one randomly selected attribute per product. For each product we take it's textual information, the attribute's value and the category. We generated synthetic versions using three strategies with controlled probabilities: correct attribute modification (50\%), incorrect examples (25\%), and unknown attributes (25\%). Detailed dataset statistics and category distributions are provided in Appendix~\ref{appendix:dataset_details}.

\subsection{Human Evaluation}
We conducted a comprehensive evaluation of our synthetic products using expert human annotators with e-commerce domain experience.  Our evaluation is exhaustive, covering all three generation strategies: correct attribute modifications, incorrect attribute examples, and unknown attribute cases. We collected three labels per product by three independent annotators, with final decisions made by majority voting. Each annotator was shown pairs of original and synthetic products with highlighted modifications (detailed protocol in Appendix~\ref{appendix:mturk}) and asked to evaluate six key aspects:

\begin{enumerate}
    \item \textbf{Attribute Value Quality}: Appropriateness of the generated value for the product category and attribute type
    
    \item \textbf{Negative Example Coherence}: For negative cases, plausibility of both the correct and incorrect values in the context
    
    \item \textbf{Cross-field Consistency}: Coherent representation of the attribute value across title, description, and features
    
    \item \textbf{Brand Modification}: Effectiveness of brand name replacements and consistency throughout the text
    
    \item \textbf{Content Preservation}: Assessment of any unintended changes beyond the target modifications
    
    \item \textbf{Professional Writing}: Overall text quality and maintenance of e-commerce writing standards
\end{enumerate}

\subsection{Downstream Task Evaluation}
To evaluate our synthetic data quality, we assess its effectiveness in attribute value extraction, a crucial e-commerce task. We focus on correct examples to enable direct comparison with the original dataset, comparing models trained on various combinations of synthetic and original data. Results demonstrate that our generated examples match the quality of real training data.

\noindent\textbf{Models and Input Processing.} We implemented attribute extraction using FLAN-T5-base\footnote{FLAN-T5 is available at \url{https://huggingface.co/google/flan-t5-base} under the Apache 2.0 license.} \cite{10.5555/3722577.3722647}, processing the product's text fields (title, description and features) in a structured prompt template with a 512-token context window. The task is framed as a generation problem, allowing for free-form attribute value prediction.

\noindent\textbf{Dataset Configuration.} From our sampled products, we utilized correct examples ($\sim$800 products) for training (80\%) and validation (20\%), evaluating on the remaining products ($\sim$1,000). We tested six configurations: zero-shot evaluation, original data only, synthetic data only, and three hybrid configurations combining original and synthetic data in different proportions (75/25, 50/50, and 25/75 splits).

\noindent\textbf{Training Setup.} We fine-tuned the model using AdamW optimizer (learning rate=5e-5) for up to 12 epochs with early stopping on validation accuracy. For generation, we used beam search (beams=5) with controlled sampling (temperature=0.7, top\_k=50, top\_p=0.95) and length constraints (max\_length=20) to ensure diverse yet accurate outputs.

%% file: sections/5_results_analysis.tex
\subsection{Human Evaluation}

\begin{table}[t]
\centering
\small
\begin{tabular}{lr}
\hline
\multicolumn{2}{c}{\textbf{Quality Metrics (\%)}} \\
\hline
\textbf{Aspect} & \textbf{Score} \\
\hline
Attribute Value Correctness & 96.5 \\
Synthetic Product Readability & 99.6 \\
Brand Modification Success & 95.8 \\
\hline
\multicolumn{2}{c}{\textbf{Consistency by Input Label (\%)}} \\
\hline
Correct & 94.2 \\
Incorrect & 93.0 \\
Unknown & 88.3 \\
\hline
\multicolumn{2}{c}{\textbf{Additional Changes (\%)}} \\
\hline
None & 88.8 \\
Acceptable & 7.0 \\
Major & 4.2 \\
\hline
\multicolumn{2}{c}{\textbf{Input Distribution (\%)}} \\
\hline
Correct & 52.0 \\
Incorrect & 24.1 \\
Unknown & 23.9 \\
\hline
\end{tabular}
\caption{Human evaluation results (N=2000 samples).}
\label{tab:human_evaluation}
\end{table}

Our human evaluation of 2,000 synthetic products demonstrates strong performance across key quality metrics (Table~\ref{tab:human_evaluation}). Expert reviewers rated 99.6\% of products as having natural e-commerce language and 96.5\% as containing valid attribute values. Brand anonymization was correctly implemented in 95.8\% of cases. Attribute consistency was maintained across all generation types: 94.2\% of products remained consistent when we modified their correct attributes, 93.0\% when we deliberately introduced incorrect values, and 88.3\% when we strategically removed attributes from the text fields. The evaluation showed that 88.8\% of products preserved their original structure completely, with 7.0\% showing minor acceptable changes and 4.2\% requiring significant revisions, primarily in cases with empty descriptions. Quantitative analysis of edge cases showed that the model successfully generated appropriate content for all empty descriptions (100\% of 47 cases) and correctly anonymized brand names in 95.8\% of products without explicit instructions. Manual analysis identified that validity and consistency errors primarily originated from issues in the source dataset, specifically overly generic attribute values (e.g., "type") and semantically misaligned attributes. A manual inspection of a small sample of the synthetic data hints to the model's ability of generating more precise and standardized attribute values compared to the original data. Future improvements could focus on implementing clearer attribute definitions, enhanced handling of semantically misaligned values, and additional validation prompts for brand handling.

\subsection{Impact on Attribute Extraction Models}
Our experiments demonstrate that synthetic data has comparable performance on real data in attribute extraction (Table~\ref{tab:finetuning_results}). Training with synthetic data alone achieves 60.48\% accuracy on our test set, performing on par with the original data (60.79\%) and substantially improving upon the zero-shot baseline of 13.4\%. This comparable performance indicates our synthetic data captures the essential patterns without introducing noise. Hybrid configurations yield even better results, with the 75\%-25\% original-synthetic mix reaching 68.82\% accuracy. Performance gradually decreases by 2.71 and 4.38 percentage points as we increase the synthetic portion to 50\% and 75\% respectively, suggesting an optimal mixing ratio. Manual review of the test predictions initially marked as incorrect (441 out of 959 examples for the synthetic only configuration) revealed that many model outputs were semantically correct but differed from MAVE's annotations in format. We identified seven types of valid variations: (1) granularity differences (e.g., ``running shoe'' vs ``running''), (2) morphological variations (``wall sticker'' vs ``wall stickers''), (3) multiple valid values (``t-shirt'' or ``tank top'' for the same product), (4) missing units (``1200'' vs ``1200 thread count''), (5) equivalent attribute definitions (``type=processor'' vs ``type=food processor''), (6) contextual synonyms (``striped'' vs ``stripe''), and (7) format variations (``ipod touch'' vs ``for apple ipod''). These findings highlight both the quality of our synthetic data generation pipeline and the inherent complexity of maintaining consistent attribute value annotations in product catalogs.

\begin{table}[t]
\centering
\small
\begin{tabular}{lc}
\toprule
\textbf{Configuration} & \textbf{Test Acc.} \\
\midrule
Zero-shot & 13.40 \\
Original-100\% & 60.79 \\
Synthetic-100\% & 60.48 \\
Original-75\% + Synthetic-25\% & 68.82 \\
Original-50\% + Synthetic-50\% & 66.11 \\
Original-25\% + Synthetic-75\% & 64.44 \\
\bottomrule
\end{tabular}
\caption{Attribute extraction results with FLAN-T5-base. Accuracy (\%) after manual validation.}
\label{tab:finetuning_results}
\end{table}

%% file: sections/6_conclusions.tex
We presented a novel approach to generate synthetic e-commerce product data at scale using LLMs, using existing products as reference templates. Expert evaluation of 2,000 synthetic products demonstrates high effectiveness: 99.6\% natural-sounding content, 96.5\% valid attribute values, and over 90\% consistent attribute usage. In downstream attribute extraction, our synthetic data matches the performance of real training data (60.5\% vs 60.8\% accuracy) while significantly improving upon the 13.4\% zero-shot baseline, with hybrid approaches reaching 68.8\% accuracy. Our framework enables rapid generation of high-quality training data that has comparable performance to real data, providing a scalable solution for bootstrapping e-commerce services when labeled data is scarce.